\relax
\documentclass[letterpaper]{article} 
\usepackage{aaai21}  
\usepackage{times}  
\usepackage{helvet} 
\usepackage{courier}  
\usepackage[hyphens]{url}  
\usepackage{graphicx} 
\usepackage{algorithm}
\usepackage{algorithmic}
\usepackage{amssymb}
\usepackage{amsmath}
\usepackage[backref]{hyperref} 

\usepackage{epsfig}
\usepackage{graphicx}
\usepackage{multirow}
\usepackage{tabularx}
\usepackage{booktabs}
\usepackage{url}
\usepackage{makecell}

\usepackage{float}
\usepackage{subfigure}
\usepackage{graphicx}

\usepackage{natbib}  
\usepackage{caption} 
\title{PP-OCRv2: Bag of Tricks for Ultra Lightweight OCR System}
\author {
    Yuning Du, Chenxia Li, Ruoyu Guo, Cheng Cui, Weiwei Liu, Jun Zhou, \\
    Bin Lu, Yehua Yang, Qiwen Liu, Xiaoguang Hu, Dianhai Yu, Yanjun Ma \\
}
\affiliations {
    Baidu Inc. \\
    \{duyuning, yangyehua\}@baidu.com
}

\begin{document}

\maketitle

\begin{abstract}
Optical Character Recognition (OCR) systems have been widely used in various of application scenarios. Designing an OCR system is still a challenging task. In previous work, we proposed a practical ultra lightweight OCR system (PP-OCR) to balance the accuracy against the efficiency. In order to improve the accuracy of PP-OCR and keep high efficiency, in this paper, we propose a more robust OCR system, i.e. PP-OCRv2. We introduce bag of tricks to train a better text detector and a better text recognizer, which include Collaborative Mutual Learning (CML),  CopyPaste, Lightweight CPU Network (PP-LCNet), Unified-Deep Mutual Learning (U-DML) and Enhanced CTCLoss. Experiments on real data show that the precision of PP-OCRv2 is 7\% higher than PP-OCR under the same inference cost. It is also comparable to the server models of the PP-OCR which uses ResNet series as backbones. All of the above mentioned models are open-sourced and the code is available in the GitHub repository PaddleOCR \footnote{https://github.com/PaddlePaddle/PaddleOCR} which is powered by PaddlePaddle \footnote{https://github.com/PaddlePaddle}.
\end{abstract}

\section{Introduction}
\begin{figure}[t]
\centering
\subfigure{
\begin{minipage}[t]{0.9\linewidth}
\centering
\includegraphics[width=\columnwidth]{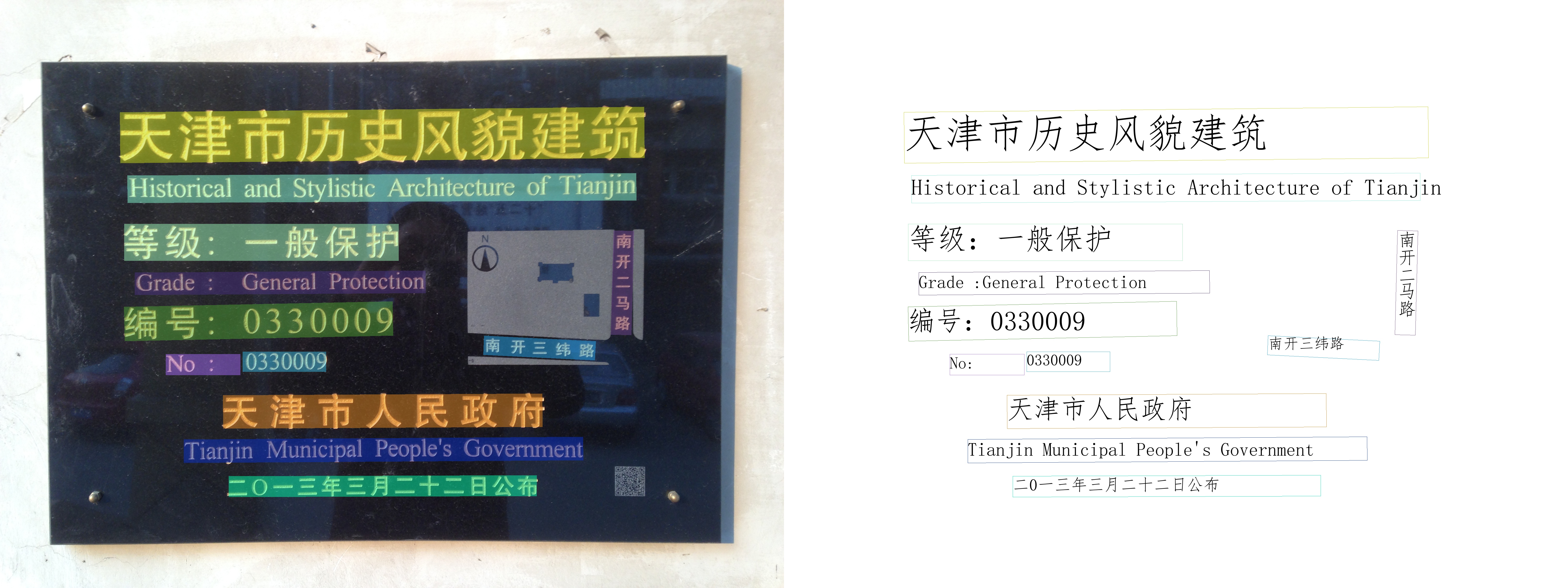}
\end{minipage}
}

\subfigure{
\begin{minipage}[t]{0.9\linewidth}
\centering
\includegraphics[width=\columnwidth]{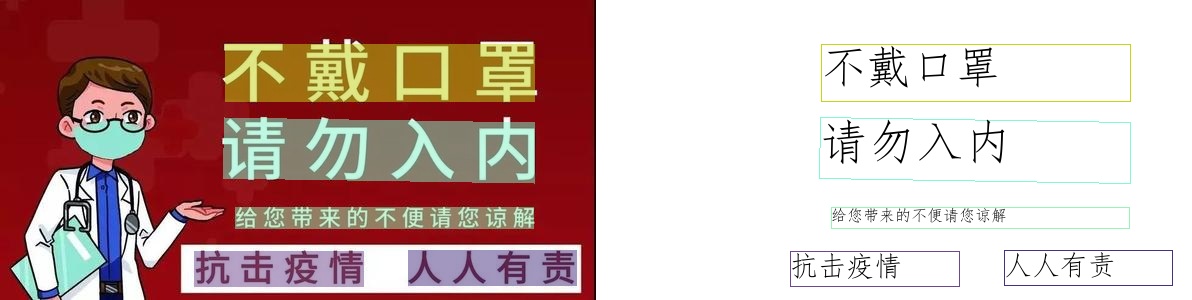}
\end{minipage}
}

\caption{Some recognition results of the proposed PP-OCRv2 system}
\label{res_p1}
\end{figure}

OCR (Optical Character Recognition) in the wild, as shown in Figure \ref{res_p1}, is well-studied in the last two decades and has various applications scenarios, such as document electronization, identity authentication, digital financial system, and vehicle license plate recognition.

\begin{figure*}[t]
\centering
\includegraphics[width=15cm]{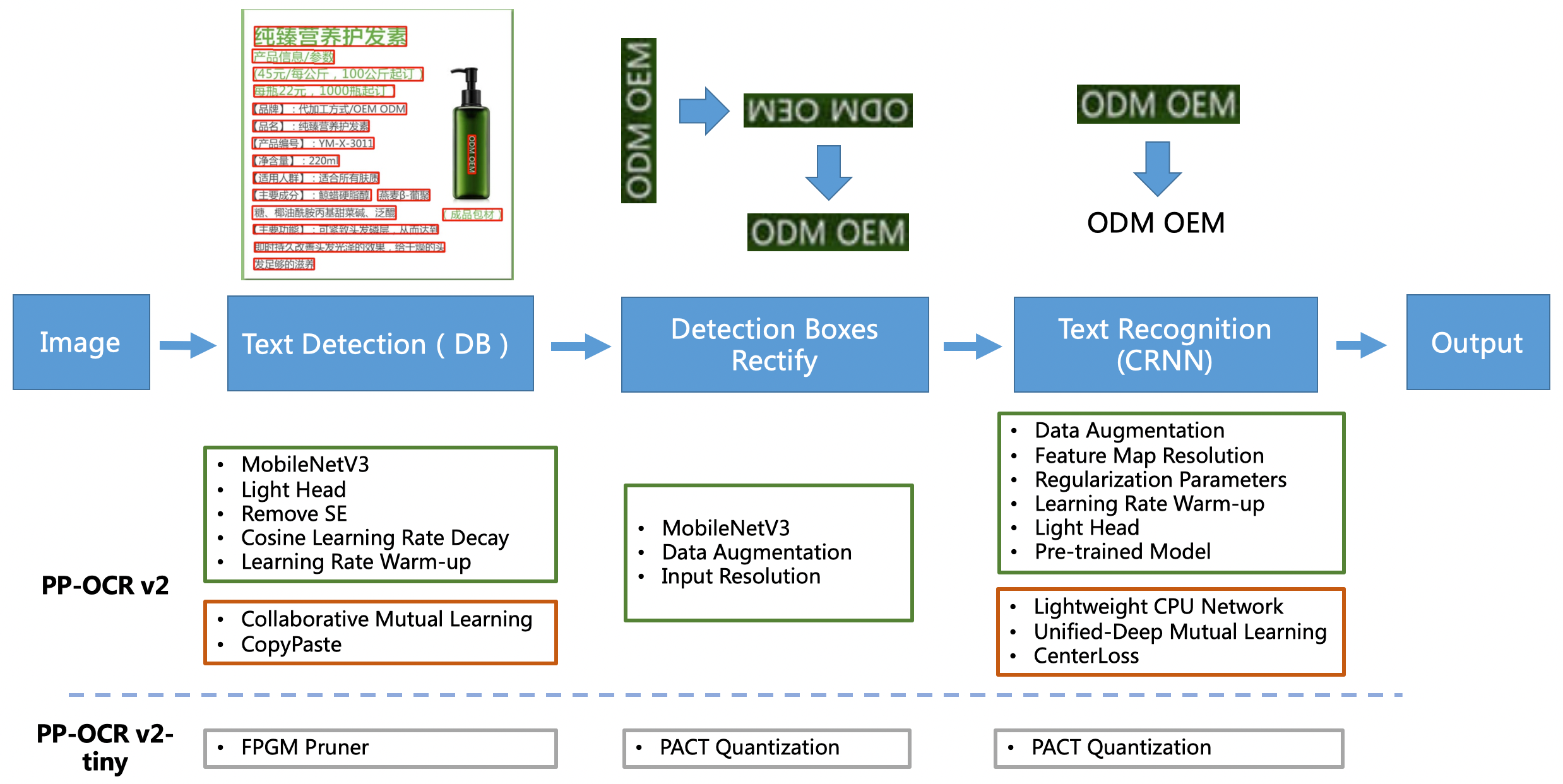}
\caption{The framework of the proposed PP-OCRv2. The strategies in the green boxes are the same as PP-OCR. The strategies in the orange boxes are the newly added ones in the PP-OCRv2. The strategies in the gray boxes will be adopted by the PP-OCRv2-tiny in the future.}
\label{framework}
\end{figure*}

When we build an OCR system in practical, not only the accuracy is considered, but also the computational efficiency. In previous, we proposed a practical ultra lightweight OCR system (PP-OCR) \cite{du2020pp} to balance the accuracy against the efficiency. It consists of three parts, text detection, detected boxes rectification and text recognition. Differentiable Binarization (DB) \cite{liao2020real} is used in text detection and CRNN \cite{shi2016end} is used in text recognition. The system adopts 19 effective strategies to optimize and slim down the size of the models. In order to improve the accuracy of the PP-OCR and keep efficiency, in this paper, we propose a more robust OCR system, i.e. PP-OCRv2. It introduces bag of tricks to train a better text detector and a better text recognizer.

Figure \ref{framework} illustrates the framework of PP-OCRv2. Most strategies follow PP-OCR as shown in the green boxes. The strategies in the orange boxes are the additional ones in PP-OCRv2. In text detection, Collaborative Mutual Learning (CML) and CopyPaste are introduced. CML utilizes two student networks and a teacher network to learn a more robust text detector. CopyPaste is a novel data augmentation trick that has been proved effectively boost performance of object detection and instance segmentation tasks \cite{ghiasi2021simple}. We show that it also works well for text detection task. In text recognition, Lightweight CPU Network (PP-LCNet)\cite{cui2021pplcnet}, Unified-Deep Mutual Learning (U-DML) and CenterLoss are introduced. PP-LCNet is a newly designed lightweight backbone based on Intel CPUs which is modified from MobileNetV1\cite{1704.04861}. U-DML utilizes two student networks to learn a more accurate text recognizer. The role of the CenterLoss is to relax the mistakes of the similar characters. We conduct a series of ablation experiments to verify the effectiveness of the above strategies.

Besides, the strategies in the gray boxes of Figure \ref{framework} are demonstrated to be effective in PP-OCR. But those are not validated in this paper. In the future, we will adopt them to speed up the inference in PP-OCRv2-tiny. 

The rest of the paper is organized as follows. In section 2, we present the details of the newly added enhancement strategies. Experimental results are discussed in section 3 and conclusions are conducted in section 4.

\section{Enhancement Strategies}

\subsection{Text Detection}
\subsubsection{Collaborative Mutual Learning (CML)}
We propose the CML method \cite{dml2017} to solve the problem of text detection distillation. There are two problems with distillation:
1. If the accuracy of the teacher model is close to that of the student model, the improvement brought by the general distillation method is limited.
2. If the structure of the teacher model and the structure of the student model are quite different, the improvement brought by the general distillation method is also very limited.

\begin{figure*}[h!]
\centering
\subfigure{
\centering
\includegraphics[width=16cm]{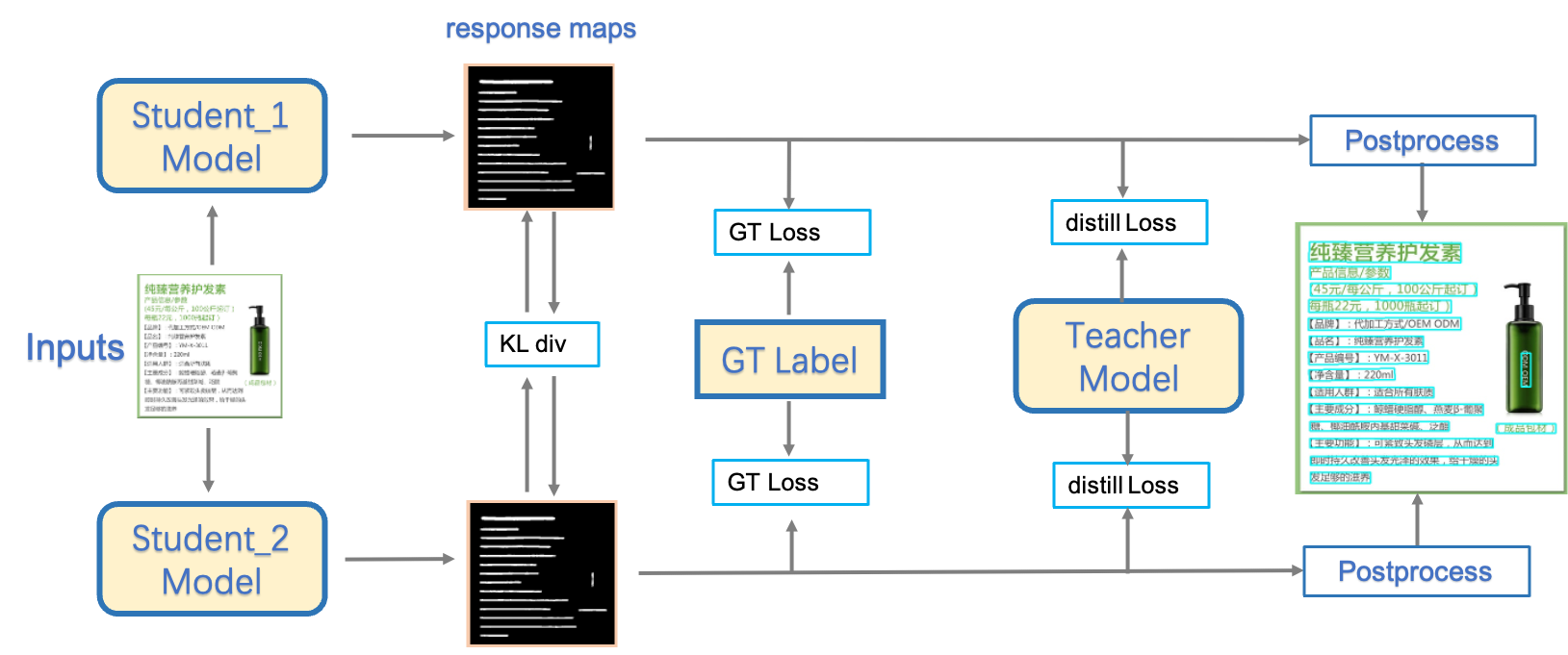}
}
\caption{The framework of CML}
\label{cml_framework}
\end{figure*}

The framework is super network composed of multiple model named student models and teacher models respectively, as illustrated in \ref{cml_framework}. And the CML method can achieve the performance that the accuracy of the student after distillation exceeds the accuracy of the teacher model in text detection. 

In CML, two sub-student models learn from each other using DML method \cite{dml2017}. Meanwhile, there is a teacher model to guide the learning of two student models. The teacher model uses ResNet18 as the backbone, and the student model uses MobinenetV3 large model with scale 0.5 as the backbone. 

CML aims to optimize sub-student model. The parameters of teacher model are freezed and only sub-student model is trained with designed losses. 
In general, the supervised information of the sub-student model has three parts, including the ground truth label, the posterior entropy of another student model and output of the teacher model. Correspondingly, there are three loss functions including the ground truth loss $L_{gt}$, peer loss from student model $L_{s}$ and distill loss from teacher model $L_{t}$.

The ground truth loss, referred as GTLoss, is to make sure that the training is supervised by the true label. We use the DB algorithm \cite{DB} to train the sub-student models. Therefore, the ground truth loss $L_{gt}$ is a combined loss, which consists of the loss of the probability map $l_{p}$, the loss of the binary map $l_{b}$, and the loss of the threshold map $l_{t}$ following DB.The formula of GTLoss is as follows, where $l_{p}$, $l_{b}$ and $l_{t}$ are binary cross-entropy loss, Dice loss  and L1 loss respectively. $\alpha $, $\beta$ are the super-parameters with default values 5 and 10 respectively.  
\begin{small}
\begin{equation}
Loss_{gt}(T_{out}, gt) = l_{p}(S_{out}, gt) + \alpha l_{b}(S_{out}, gt) + \beta l_{t}(S_{out}, gt)
\end{equation}
\label{loss_dml}
\end{small}

The sub-student models learn from each other with reference to DML method \cite{dml2017}. But the difference with DML is that the sub-student models is trained simultaneously in every iteration to speed up the training process. KL divergence is used to compute the distance between student models. The peer loss between student models is as follows, 
\begin{small}
\begin{equation}
Loss_{dml} = \frac{KL(S1_{pout} || S2_{pout}) + KL(S2_{pout} || S1_{pout})}{2}
\end{equation}
\label{loss_dml}
\end{small}

The distill loss reflects the supervision of the teacher model on the sub-student models. Teacher model can provide a wealth of knowledge to student models which is important for performance improvement.
To get better knowledge, we dilate the response probability maps of the teacher model to increase the object area. This operation can slightly improve the accuracy of the teacher model. The distillation loss is as follows, where $l_{p}$, $l_{b}$  are binary cross-entropy loss and Dice Loss respectively. And $ \gamma $ is the super-parameter default as 5. The $f_{dila}$ is the dilation function which kernel is matrix $[[1,1],[1,1]]$.
\begin{small}
\begin{equation}
Loss_{distill} = \gamma l_{p}(S_{out}, f_{dila}(T_{out})) + l_{b}(S_{out}, f_{dila}(T_{out}))
\end{equation}
\label{loss_dml}
\end{small}

Finally, the loss function used in the CML method of training the PP-OCR detection model is as follows.
\begin{small}
\begin{equation}
Loss_{total} = Loss_{gt} + Loss_{dml} + Loss_{distill} 
\end{equation}
\label{loss_dml}
\end{small}

\subsubsection{CopyPaste} 
\begin{figure}[h!]
\centering
\subfigure{
\centering
\includegraphics[width=\columnwidth]{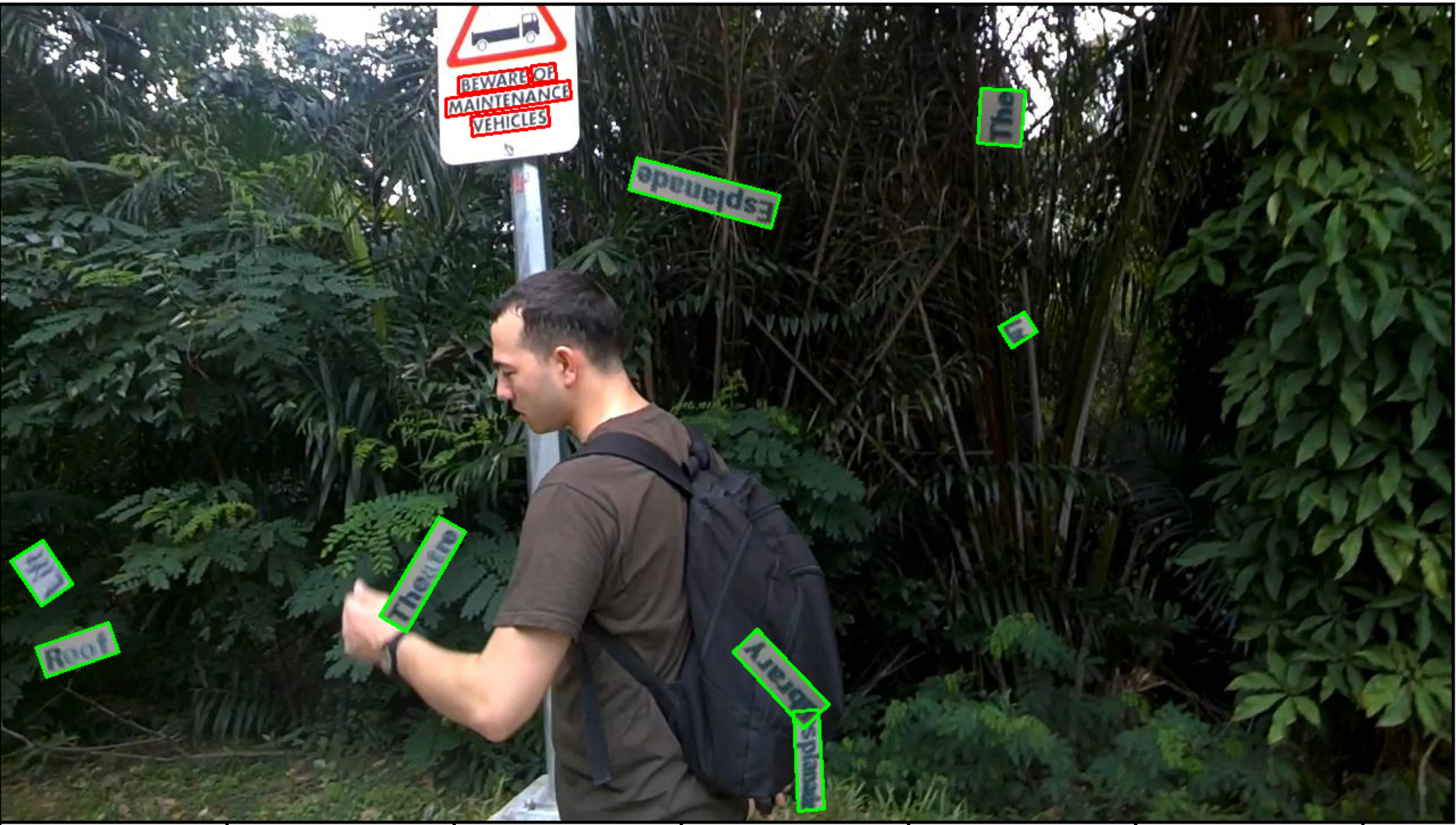}
}
\caption{Example of CopyPaste in text detection. The green boxes are pasted text, the red boxes are the original texts in the image.}
\label{CopyPaste}
\end{figure}

is a novel data augmentation trick that has been proved to be effective in boosting performance of object detection and instance segmentation tasks \cite{ghiasi2021simple}. It can synthesize text instances to balance the ratio between positive and negative samples in the training set, which traditional image rotation, random flip and random cropping cannot achieve. Due to all texts in the foreground being independent, CopyPaste pastes texts without overlapping on a randomly selected background image. Figure \ref{CopyPaste} is an example of CopyPaste.

\subsection{Text Recognition}
\subsubsection{Lightweight CPU Network (PP-LCNet)}

\begin{figure}[h!]
\centering
\subfigure{
\centering
\includegraphics[width=\columnwidth]{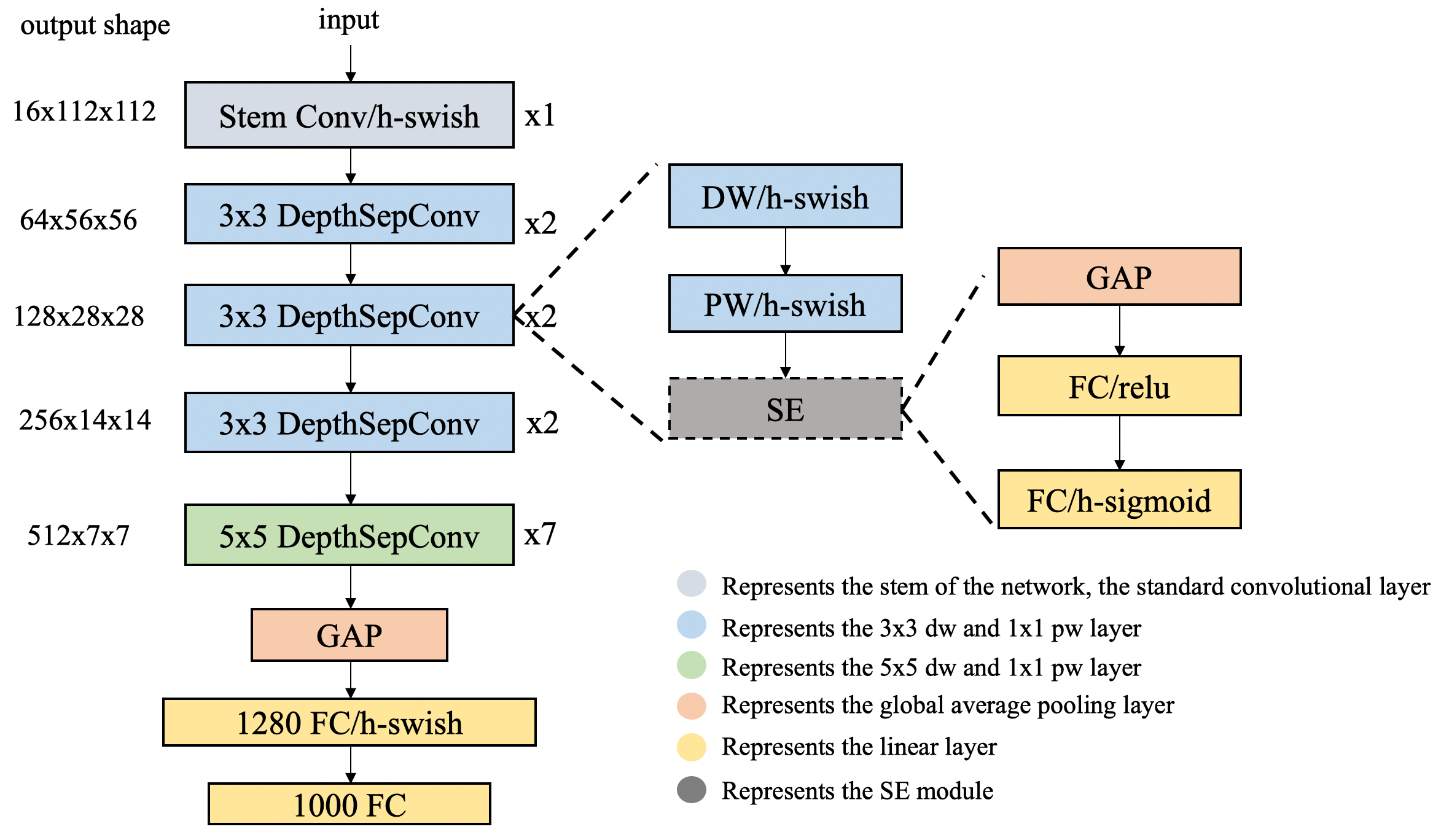}
}
\caption{PP-LCNet network structure. The dotted box represents optional modules. The stem part uses standard convolution.  DepthSepConv means depthwise separable convolutions, DW means depthwise convolutions, PW means pointwise convolutions, GAP means global average pooling.}
\label{PP-LCNet}
\end{figure}

In order to get a better accuracy-speed trade-off on Intel CPU, we have designed a lightweight backbone based on Intel CPUs, which provides a faster and more accurate OCR recognition algorithm with mkldnn enabled. The structure of the entire network is shown in Figure \ref{PP-LCNet}. Compared with MobileNetV3, as the structure of MobileNetV1 makes it easier to optimize the inference speed when MKLDNN is enabled on Intel CPU, so the network is based on MobileNetV1\cite{1704.04861}.In order to make MobileNetV1 have a stronger ability to extract features, we have made some changes to its network structure. The improvement strategy will be explained in the following four aspects.

\textit{1. Better activation function.}
In order to increase the fitting ability of MobileNetV1, we replaced the activation function in the network with H-Swish from the original ReLU, which can bring a significant improvement in accuracy with only a slight increase in inference time.

\textit{2. SE modules at appropriate positions.}
SE \cite{Hu_2018_CVPR} module has been used by a large number of networks since its being proposed. It is a good way to weight the network channels to obtain better features, and is used in many lightweight networks such as MobileNetV3\cite{Howard_2019_ICCV}. However, on Intel CPUs, the SE module increases the inference time, so that we cannot use it for the whole network. In fact, through extensive experiments, we have found that the closer to the tail of the network, the more effective the SE module is. So we just add the SE module to the blocks near the tail of the network. This results in a better accuracy-speed balance. The activation functions of the two layers in the SE module are ReLU and H-Sigmoid respectively.

\textit{3. Larger convolution kernels.}
The size of the convolution kernel often affects the final performance of the network. In mixnet\cite{1907.09595}, the authors analyzed the effect of differently sized convolution kernels on the performance of the network, and finally mixed differently sized kernels in the same layer of the network. However, such a mixture slows down the inference speed of the model, so we tried to increase the size of the convolution kernels with as little increase in inference time as possible. In the end, we set the size of the convolution kernel at the tail of the network as $5 \times 5$.

\textit{4. Larger dimensional 1x1 conv layer after GAP.}
In PP-LCNet, the output dimension of the network after GAP is small, and directly connecting the final classification layer will lose the combination of features. In order to give the network a stronger fitting ability, we connected a 1280-dimensional size 1x1 conv to the final GAP layer, which would increase the model size without increasing the inference time.

With these four changes, our model performs well on the ImageNet, and table \ref{PP-LCNet-Ablation} lists the metrics against other lightweight models on Intel CPUs.

\subsubsection{Unified-Deep Mutual Learning (U-DML)}

Deep mutual learning \cite{dml2017} is a method in which two student networks learn by each other, and a larger teacher network with pre-trained weights is not required for knowledge distillation. In DML, for image classification task, the loss functions contains two parts: (1) loss function between student networks and groundtruth. (2) Kullback–Leibler divergence (KL-Div) loss among the student networks' output soft labels.

Heo proposed OverHaul \cite{Overhaul}, in which feature map distance between student network and teacher network are used for the distillation process. Transform is carried out on student network feature map to keep the feature map alignment.

\begin{figure*}[h!]
\centering
\subfigure{
\centering
\includegraphics[width=14cm]{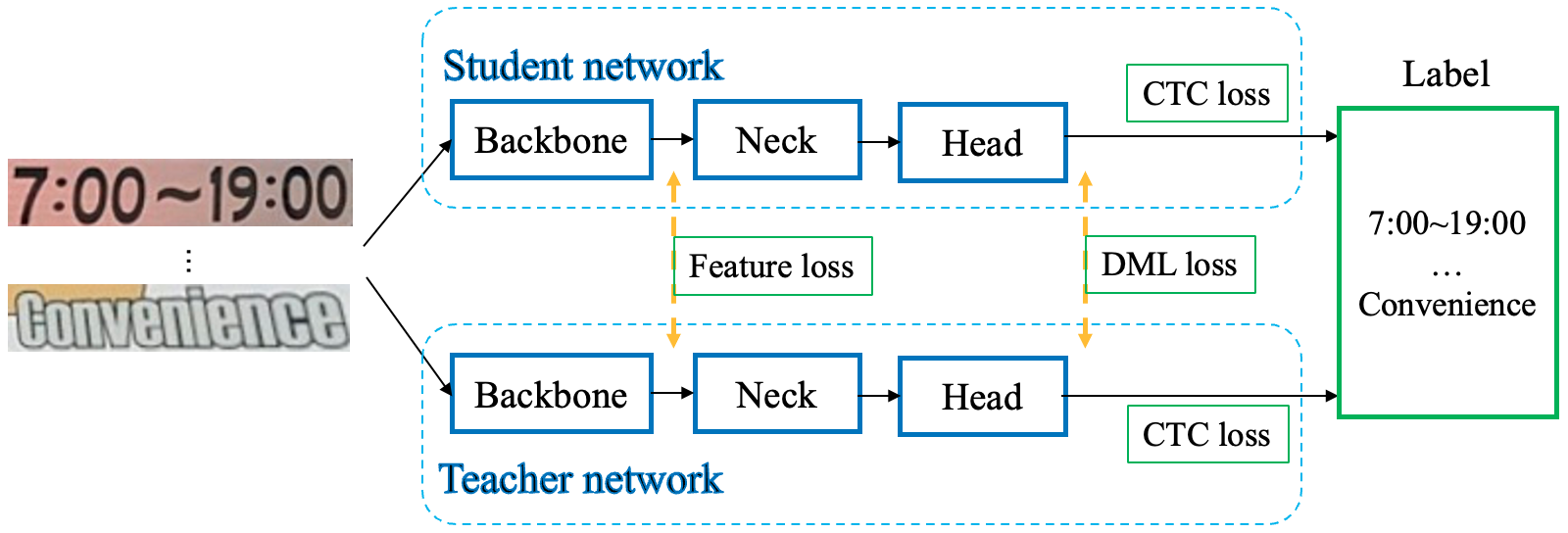}
}
\caption{U-DML framework}
\label{udml_framework}
\end{figure*}

To avoid too time-consuming teacher model training process, in this paper, based on DML, we proposed U-DML, in which feature maps are also supervised during the distillation process. Figure \ref{udml_framework} shows the framework of U-DML.

There are two networks for the distillation process: the student network and the teacher network. They have exactly the same network structures with different initialized weights. The goal is that for the same input image, the two networks can get the same output, not only for the prediction result but also for the feature map.

The total loss function consists of three parts: (1) CTC loss. Since the two networks are trained from scratch, CTC loss can be used for the networks' convergence; (2) DML loss. It's expected that two the networks' final output distributions are same, so DML loss is needed to guarantee the consistency of distribution between the two networks; (3) Feature loss. The two networks' architectures are same, so their feature maps are expected to be same, feature loss can be used to constrain the two networks' intermediate feature map distance.

\textit{CTC loss.}
CRNN is the base architecture for text recognition in this paper, which integrates feature extraction and sequence modeling. It adopts the Connectionist Temporal Classification (CTC) loss \cite{graves2006connectionist} to avoid the inconsistency between prediction and groundtruth. Since the two sub-networks are trained from scratch, CTC loss is adopted for the two sub-networks. The loss function is as follows.

\begin{small}
\begin{equation}
Loss_{ctc} = CTC(S_{hout}, gt) + CTC(T_{hout}, gt)
\end{equation}
\label{loss_ctc}
\end{small}

in which $S_{hout}$ denotes head output of the student network and $T_{hout}$ denotes that of the teacher network. $gt$ donates the groudtruth label of the input image.

\textit{DML loss.}
In DML, parameters of each sub-network are updated separately. Here, to simplify the training process, we calculate the KL divergence loss between the two sub-networks and update all the parameter simultaneously. The DML loss is as follows.

\begin{small}
\begin{equation}
Loss_{dml} = \frac{KL(S_{pout} || T_{pout}) + KL(T_{pout} || S_{pout})}{2}
\end{equation}
\label{loss_dml}
\end{small}

in which $KL(p || q)$ denotes KL divergence of the $p$ and $q$. $S_{pout}$ and $T_{pout}$ can be calculated as follows.

\begin{small}
\begin{equation}
\begin{aligned}
S_{pout} = Softmax(S_{hout}) \\
T_{pout} = Softmax(T_{hout})
\end{aligned}
\end{equation}
\label{pout_calc}
\end{small}

\textit{Feature loss.}
During the training process, we hope that the backbone output of the student network is same as that of the teacher network. Therefore, similar to Overhaul, feature loss is used for the distillation process. The loss can be calculated as follows.

\begin{small}
\begin{equation}
Loss_{feat} = L2(S_{bout}, T_{bout})
\end{equation}
\label{loss_feat}
\end{small}

in which $S_{bout}$ means backbone output of the student network and $T_{bout}$ means that of teacher network. Mean square error Loss is utilized here. It is noted that for the feature loss,  feature map transformation is not needed because the two feature maps used to calculate the loss are exactly the same.

Finally, the total loss for the U-DML training process is shown as follows.

\begin{small}
\begin{equation}
Loss_{total} = Loss_{ctc} + Loss_{dml} + Loss_{feat}
\end{equation}
\label{loss_udml_total}
\end{small}

During the training process, we find that piece-wise learning rate strategy is a better choice for distillation. When the feature loss is used, it takes a longer time for the model to reach the best accuracy, so 800 epochs and piece-wise strategy are utilized here for the text recognition distillation process.

Moreover, for the standard CRNN architecture, just one FC layer is used in CTC-Head, which is slightly weak for the information decoding process. Therefore, we modify the CTC-Head part, using two FC layers, this leads to about 1.5\% accuracy improvement without any extra inference time cost.

\subsubsection{Enhanced CTCLoss}

There exists a lot of similar characters in Chinese recognition task. Their differences in appearance are very small which are often mistakenly recognized. In PP-OCRv2, We designed an enhanced CTCLoss, which combined the original CTCLoss and the idea of CenterLoss \cite{wen2016discriminative} in metric learning. Some improvements are made to make it suitable for sequence recognition Task. Enhanced CTCLoss is defined as follows: 
\begin{small}
\begin{equation}
          L = L_{ctc} + \lambda * L_{center}
\end{equation}
\end{small}
\begin{small}
\begin{equation}
L_{center} =\sum_{t=1}^T||x_{t} - c_{y_{t}}||_{2}^{2}
\end{equation}
\end{small}in which, $x_{t}$ is the feature of timestamp $t$. $c_{y_{t}}$ is the center of class $y_{t}$. We have no explicit label $y_{t}$ for $x_{t}$ because of the misalignment between features and labels in CRNN \cite{shi2016end} algorithm. We adopt the greedy decoding strategy to get $y_{t}$:
\begin{small}
\begin{equation}
          y_{t} = argmax(W * x_{t})
\end{equation}
\end{small}$W$ is the parameters of CTC head. Experiments show that: $\lambda$ = 0.05 is a good choice.

\section{Experiments}

\subsection{Experimental Setup}

\subsubsection{DataSets}
We perform experiments on the same datasets as we used in our previous work PP-OCR \cite{du2020pp} as shown in Table \ref{dataset_ch}. 

\begin{table*}[h]
\begin{center}
\begin{tabular}{c|c|c|c|c}
\hline
& \multicolumn{3}{c|}{Number of training data} & \makecell[c]{Number of validation data} \\
\cline{2-5}
Task & Total & Real & Synthesis & Real \\
\hline
Text Detection & 97K & 68K & 29K & 500  \\
Text Recognition & 17.9M & 1.9M & 16M & 18.7K  \\
\hline
\end{tabular}
\end{center}
\caption{Statistics of dataset for text detection and recognition.}
\label{dataset_ch}
\end{table*}

For text detection, there are 97k training images and 500 validation images. The training images consist of 68K real scene images and 29K synthetic images. The real scene images are collected from  Baidu image search and public datasets include LSVT \cite{sun2019chinese}, RCTW-17 \cite{shi2017icdar2017}, MTWI 2018 \cite{mtwi}, CASIA-10K \cite{he2018multi}, SROIE \cite{huang2019icdar2019}, MLT 2019 \cite{nayef2019icdar2019}, BDI \cite{karatzas2011icdar}, MSRA-TD500 \cite{yao2012detecting} and CCPD 2019 \cite{xu2018towards}. The synthetic images mainly focus on the scenarios for long texts, multi-direction texts and texts in table. The  validation images are all from real scenes.

For text recognition, there are 17.9M training images and 18.7K validation images. Among the training images, 1.9M images are real scene images, which come from some public datasets and Baidu image search. The public datasets used include LSVT, RCTW-17, MTWI 2018 and CCPD 2019. The remaining 16M synthetic images mainly focus on scenarios for different backgrounds, rotation, perspective transformation, noising, vertical text, etc. The corpus of synthetic images comes from the real scene images. All the validation images also come from the real scenes.

\begin{figure}[h!]
\centering
\subfigure{
\centering
\includegraphics[width=\columnwidth]{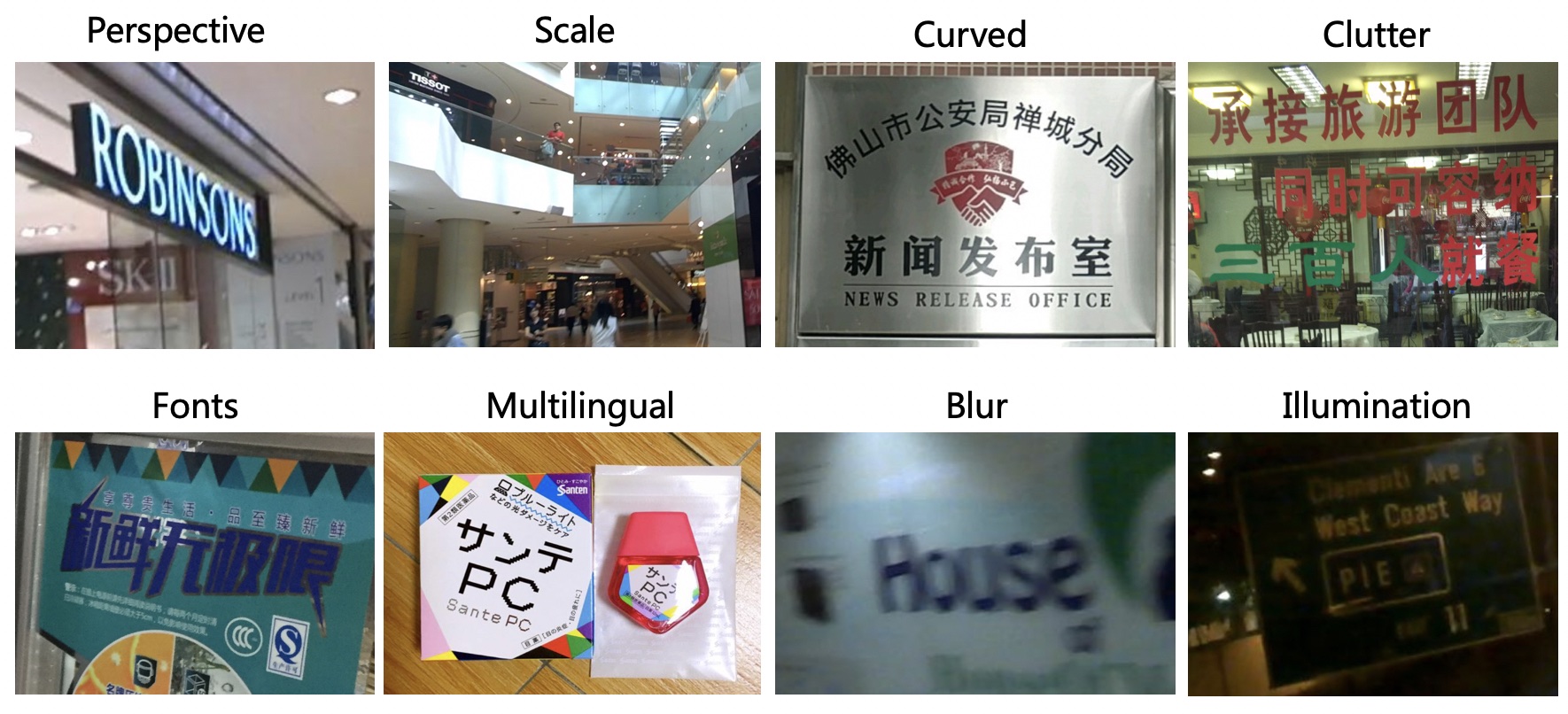}
}
\caption{Some images contained scene text}
\label{SRT}
\end{figure}

\begin{figure}[h!]
\centering
\subfigure{
\centering
\includegraphics[width=\columnwidth]{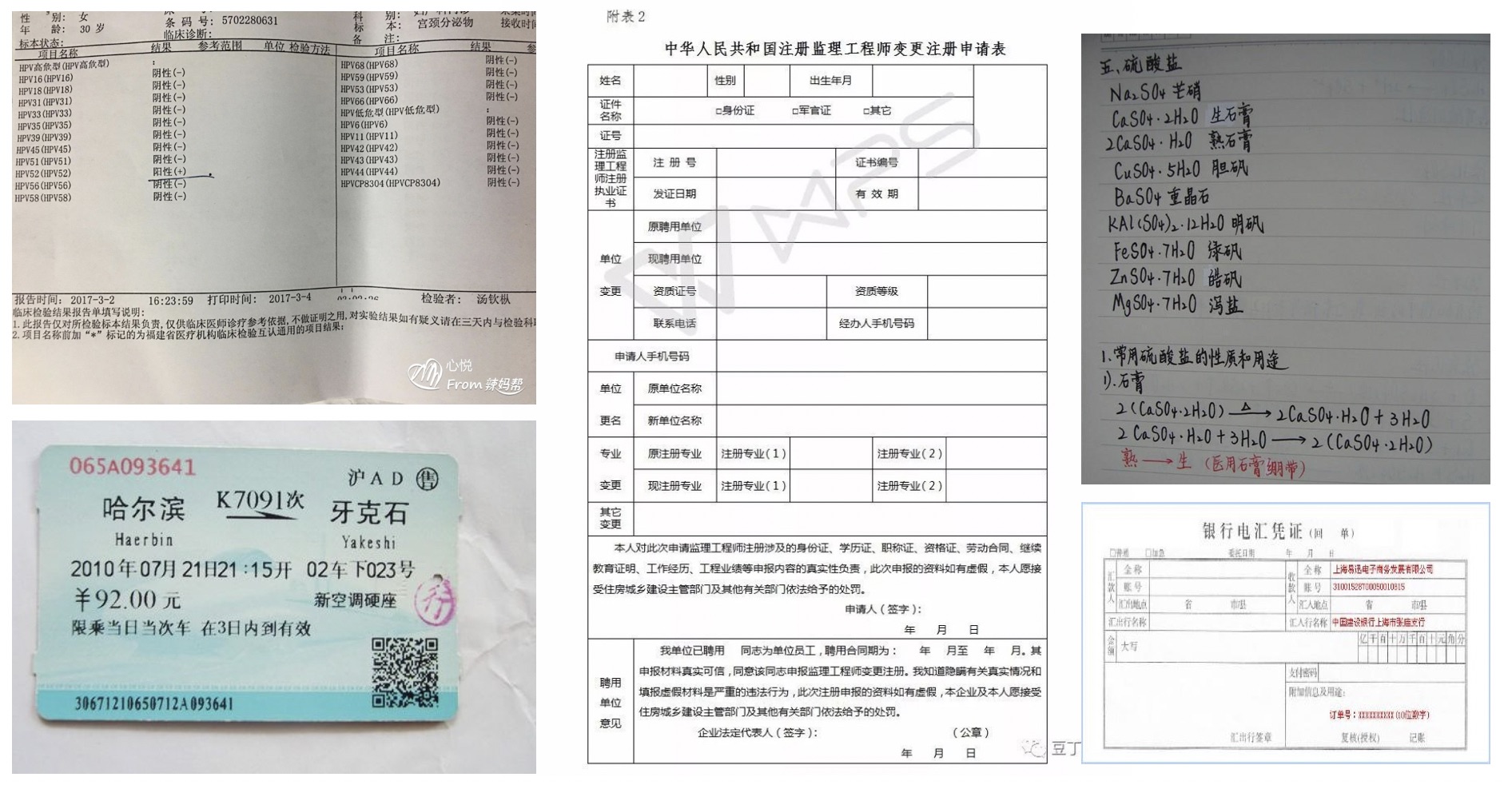}
}
\caption{Some images contained document text}
\label{doc}
\end{figure}

In addition, we collected 300 images for different real application scenarios to evaluate the overall OCR system, including contract samples, license plates, nameplates, train tickets, test sheets, forms, certificates, street view images, business cards, digital meter, etc. Figure \ref{SRT} and Figure \ref{doc} show some images of the test set.

The data synthesis tool used in text detection and text recognition is modified from text render \cite{textrender}.
    
\subsubsection{Implementation Details}
We adopt most of the strategies used in PP-OCR \cite{du2020pp} as you can found in Figure \ref{framework}. We use Adam optimizer to train all the models, setting the initial learning rate to 0.001. The difference is that we adopt cosine learning rate decay as the learning rate schedule for the training of detection model, but piece-wise decay for recognition model training. Warm-up training for a few epochs at the beginning is utilized for both detection and recognition models training.

For text detection, the model is trained for 700 epochs in total with warm-up training for 2 epochs. The batch size is set to 8 per card. For text recognition, the model warm up for 5 epochs and is then trained for 700 epochs with the initial learning rate 0.001, and then trained for 100 epochs with learning rate decayed to 0.0001. The batch size is 128 per card.

In the inference period, Hmean is used to evaluate the performance of the text detector and the end-to-end OCR system. Sentence Accuracy is used to evaluate the performance of the text recognizer. GPU inference time is tested on a single T4 GPU. CPU inference time is tested on a Intel(R) Xeon(R) Gold 6148.
    
\subsection{Text Detection}
Table \ref{det_strategy} shows the ablation study of DML, CML and CopyPaste for text detection. The baseline model is PP-OCR lightweight detection model. The long side of input image is resized to 960 during the test. As the data shows, DML can improve the Hmean metric by nearly 2\%, while CML can improve by 3\%. At last, the final Hmean can be  further improved by 0.6\% by the data augmentation method CopyPaste. So PP-OCRv2 detection model yields a 3.6\% improvement over PP-OCR at the same speed, as the model structure stays the same. The inference time is the overall time consumed including pre-processing and post-processing.

\begin{table*}[h]
\begin{center}
\begin{tabular}{l|c|c|c|c|c}
\hline
Strategy  & Precision & Recall & Hmean & \multicolumn{1}{c}{\begin{tabular}[c]{@{}c@{}}Model Size\\ (M)\end{tabular}} & \multicolumn{1}{c}{\begin{tabular}[c]{@{}c@{}}Inference Time\\ (CPU, ms)\end{tabular}} \\
\hline
PP-OCR det                   &    0.718    &  0.805    &   0.759    &    3.0    & 129 \\
PP-OCR det + DML             &    0.743    &  0.815    &   0.777    &    3.0    & 129 \\
PP-OCR det + CML             &    0.746    &  0.835    &   0.789    &    3.0    & 129 \\
PP-OCR det + CML + CopyPaste &    0.754    &  0.840    &   0.795    &    3.0    & 129 \\ 
\hline
\end{tabular}
\end{center}
\caption{Ablation study of CML and CopyPaste for text detection.}
\label{det_strategy}
\end{table*}

\subsection{Text Recognition}
Table \ref{rec_strategy} shows the ablation study of PP-LCNet, U-DML and Enhanced CTC loss. Comparing PP-LCNet with MV3, the accuracy can be improved by 2.6\%. Even though the model size with PP-LCNet is 3M bigger, the inference time is reduced from 7.7ms to 6.2ms due to the reasonable design of the network structure.  The U-DML method can improve the accuracy by another 4.6\%, which is a significant improvement. Further more, the accuracy can be improved by 0.9\% with Enhanced CTC loss. So with all these strategies, the accuracy is improved by 8.1\%, with model size 3M bigger but average inference time 1.5ms faster.

\begin{table*}[h]
\begin{center}
\begin{tabular}{l|c|c|c}
\hline
Strategy   & Acc & \multicolumn{1}{c}{\begin{tabular}[c]{@{}c@{}}Model Size\\ (M)\end{tabular}} & \multicolumn{1}{c}{\begin{tabular}[c]{@{}c@{}}Inference Time\\ (CPU, ms)\end{tabular}} \\
\hline
PP-OCR rec (MV3)      & 0.667&  5.0      &     7.7  \\
PP-OCR rec (PP-LCNet)       &   0.693   &    8.0    &     6.2  \\
PP-OCR rec (PP-LCNet) + U-DML     &  0.739    &   8.6     &      6.2  \\
PP-OCR rec (PP-LCNet) + U-DML + Enhanced CTC loss    &  0.748  & 8.6   & 6.2 \\ 
\hline
\end{tabular}
\end{center}
\caption{Ablation study of PP-LCNet, U-DML, and Enhanced CTC loss for text recognition.}
\label{rec_strategy}
\end{table*}

\subsubsection{Ablation study for PP-LCNet}
In order to test the generalization ability of PP-LCNet, we used challenging datasets like ImageNet-1k throughout the process of designing the model. Table \ref{PP-LCNet-Ablation} shows the accuracy-speed comparison among the PP-LCNet and other different lightweight models that we have selected for comparable accuracy on ImageNet. It is obvious that PP-LCNet achieves better performance from both speed and accuracy, even when compared to a very competitive network like MobileNetV3.

\begin{table}[h]
\begin{center}
\begin{tabular}{l|c|c}
\hline
Model & \shortstack{Top1-Acc\\(\%)} & \shortstack{Inference Time\\(ms)} \\
\hline
MobileNetV1-0.75x & 68.81 & 3.88 \\
MobileNetV2-0.75x & 69.83 & 4.56\\
MobileNetV3-small-1.0x & 68.24 & 4.20 \\
MobileNetV3-large-0.5x & 69.24 & 4.54\\
GhostNet-0.5x & 66.88 & 6.63\\
\textbf{PP-LCNet-1.0x} & \textbf{71.32} & \textbf{3.16}\\
\hline
\end{tabular}
\end{center}
\caption{Metrics of different lightweight models on ImageNet-1k, the CPU used in the test is Intel(R)-Xeon(R)-Gold-6148-CPU, the resolution of the image is 224x224, the batch-size is 1, the number of thread is 4, the option of MKLDNN is on, and the final inference time is the mean inference time of 30000 images.}
\label{PP-LCNet-Ablation}
\end{table}

\subsection{System Performance}
\begin{figure*}[h]
\centering

\subfigure[]{
\includegraphics[width=0.3\textwidth]{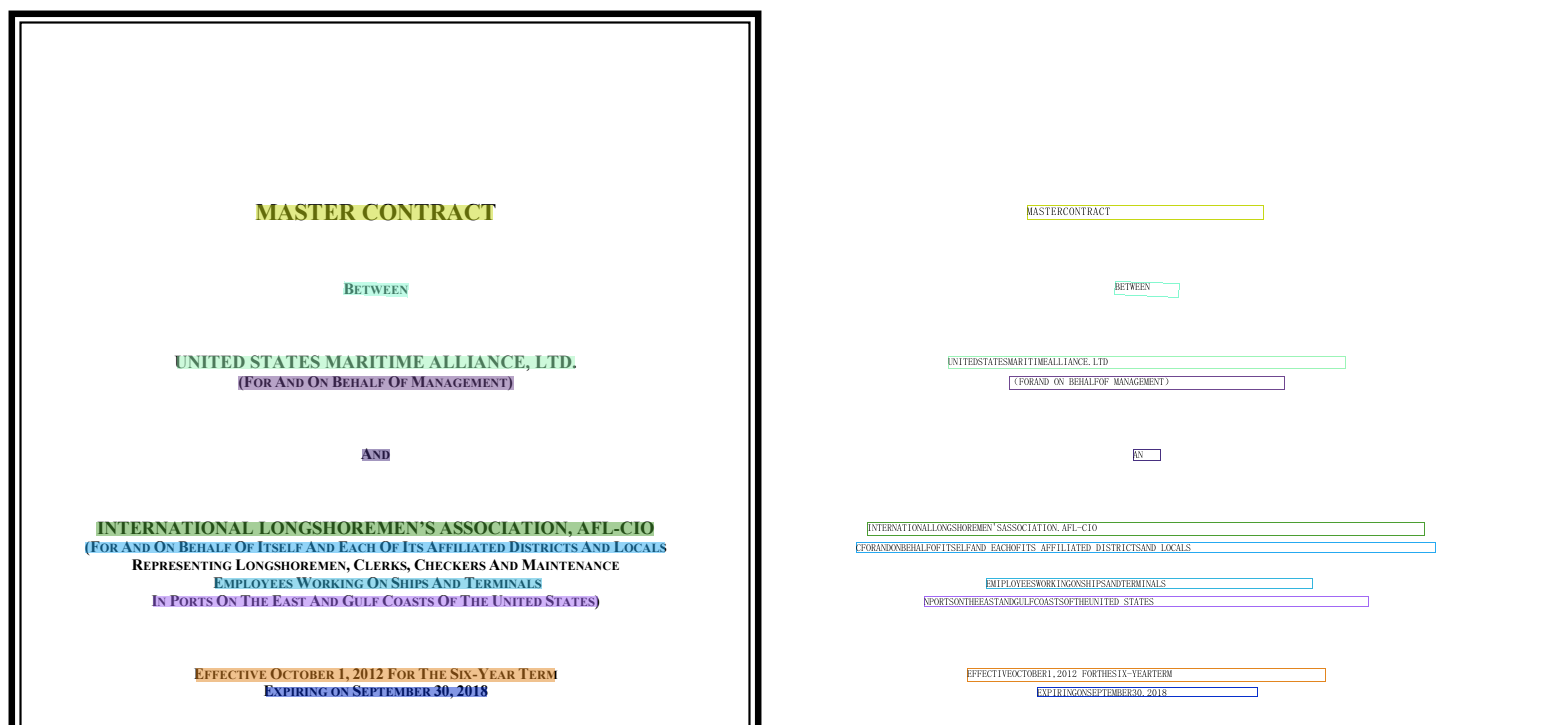}
}
\subfigure[]{
\includegraphics[width=0.3\textwidth]{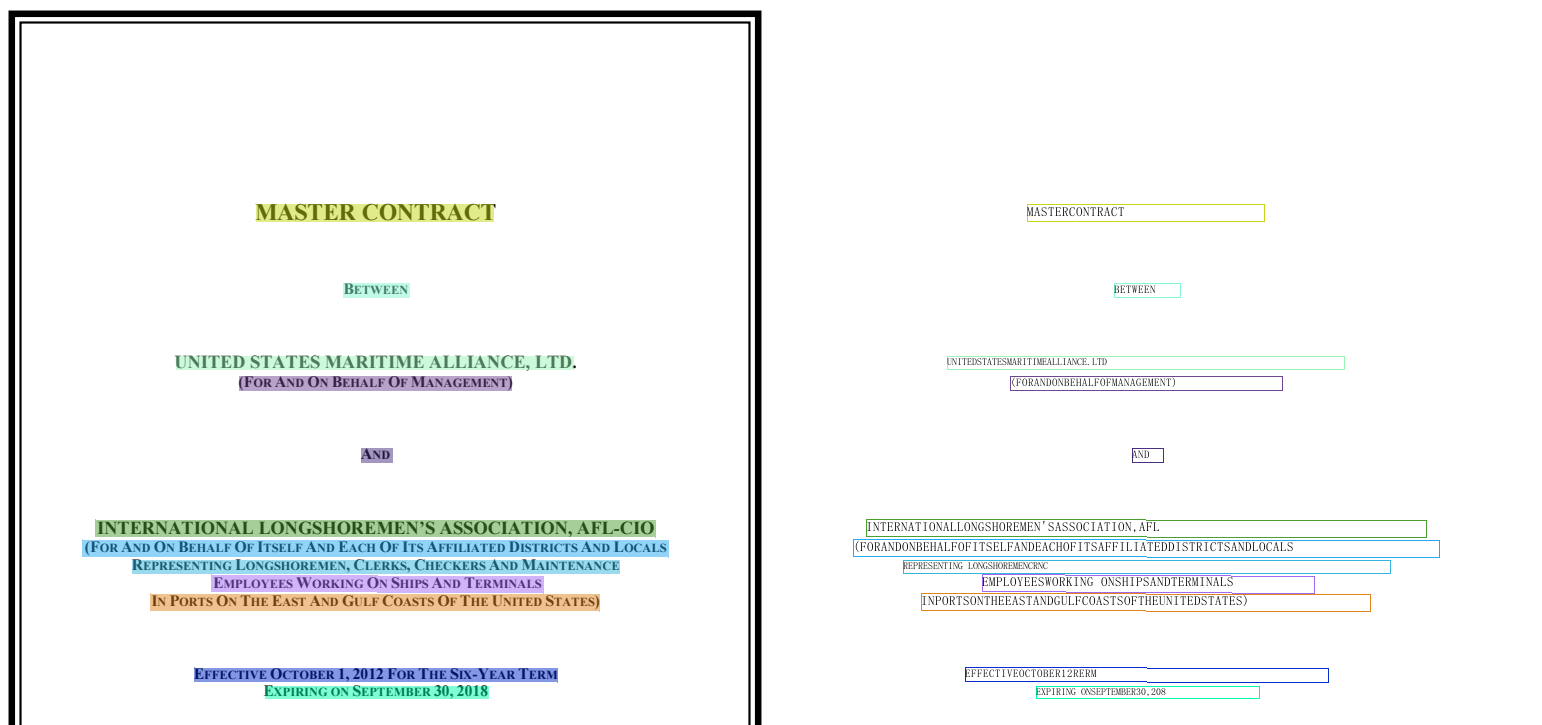}
}
\subfigure[]{
\includegraphics[width=0.3\textwidth]{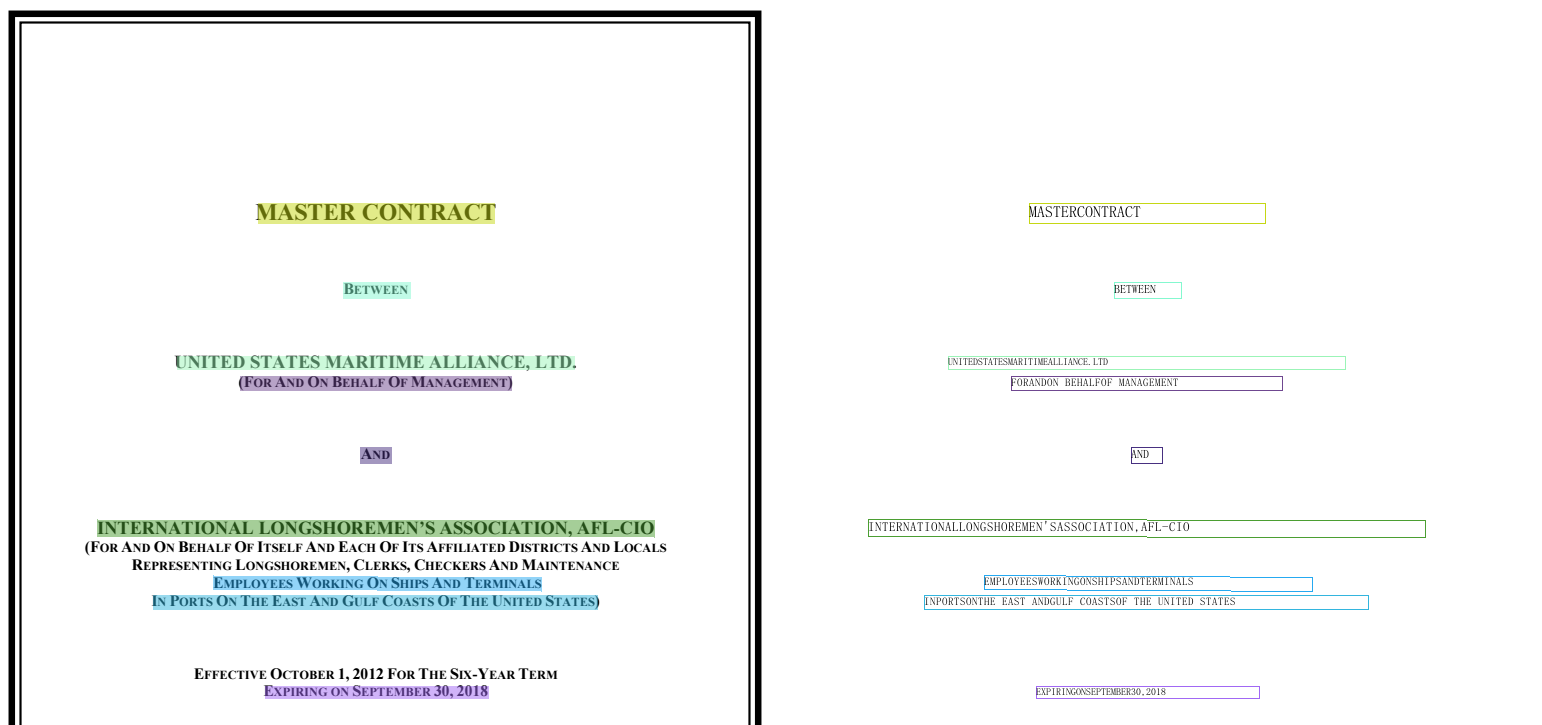}
}
\hfill
\subfigure[]{
\includegraphics[width=0.3\textwidth]{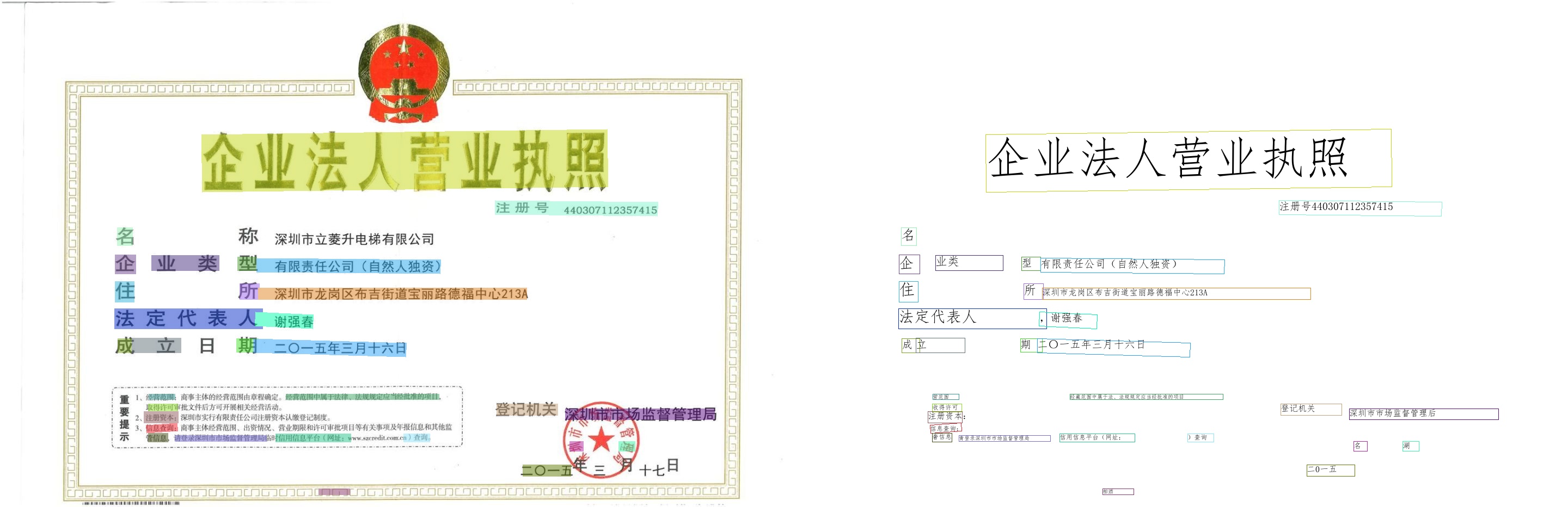}
}
\subfigure[]{
\includegraphics[width=0.3\textwidth]{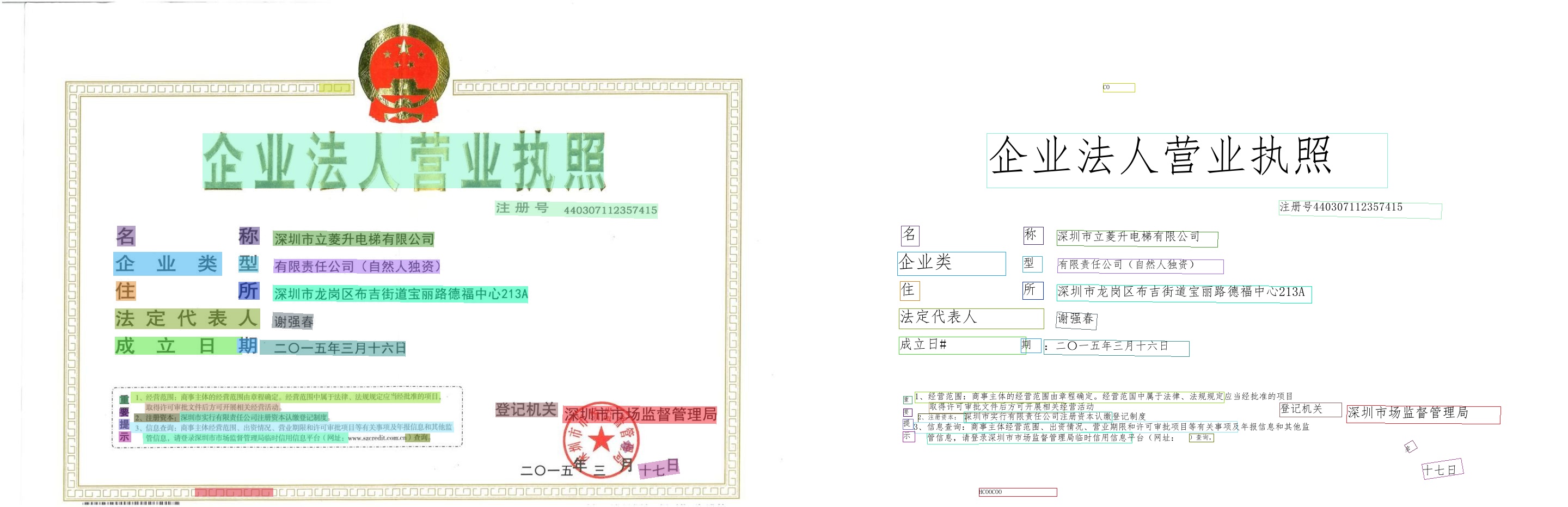}
}
\subfigure[]{
\includegraphics[width=0.3\textwidth]{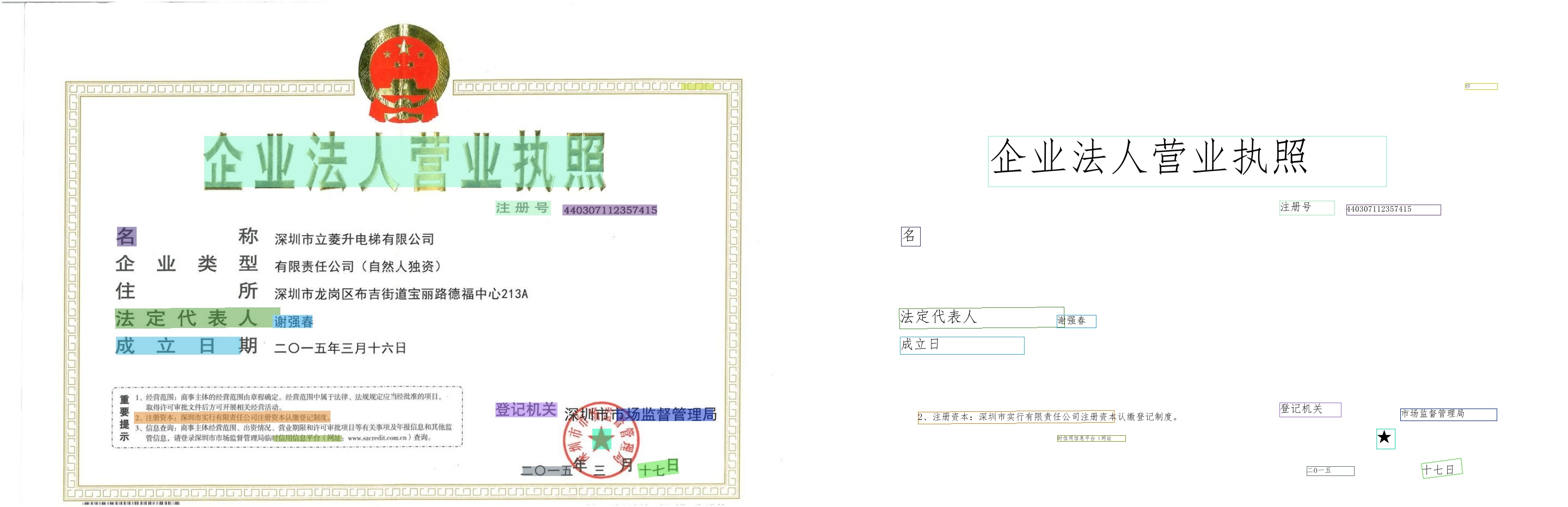}
}
\hfill
\subfigure[]{
\includegraphics[width=0.3\textwidth]{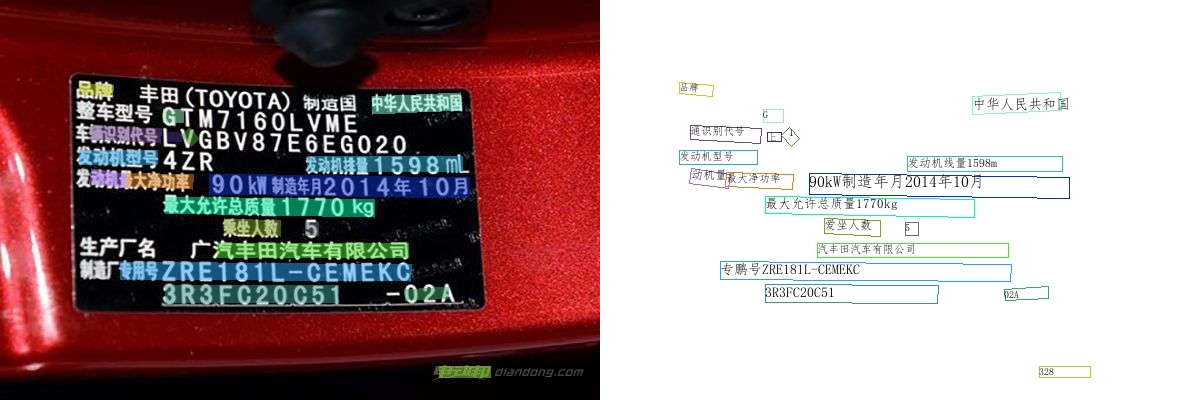}
}
\subfigure[]{
\includegraphics[width=0.3\textwidth]{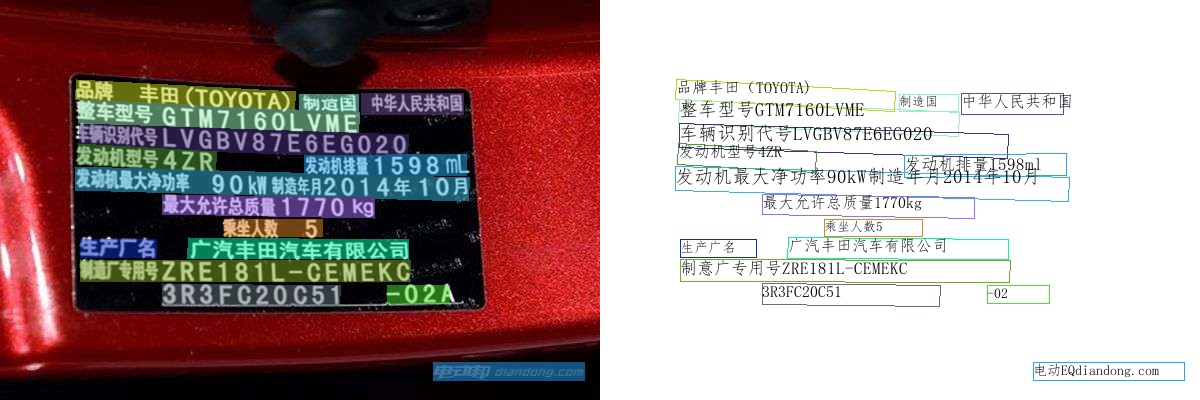}
}
\subfigure[]{
\includegraphics[width=0.3\textwidth]{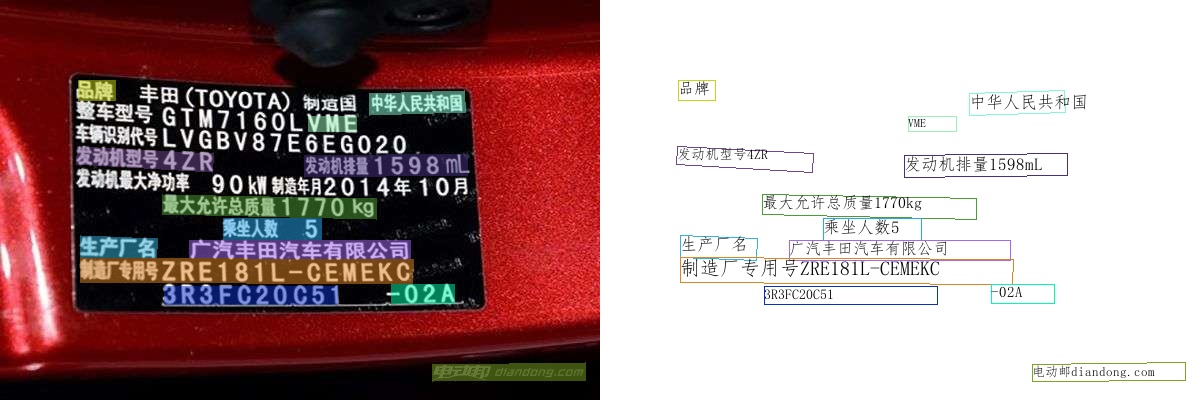}
}

\caption{Some images results of the proposed PP-OCRv2 system and the previous ultra lightweight and large-scale PP-OCR system. (a)(d)(g) results of PP-OCR mobile system. (b)(e)(h) results of PP-OCRv2 system. (c)(f)(i) results of PP-OCR  server system.}
\label{show_res}
\end{figure*}

In Table \ref{sys_scale}, we compare the performance between proposed PP-OCRv2 with the previous ultra lightweight and large-scale PP-OCR system. The large-scale PP-OCR system, which uses ResNet18\_vd as the text detector backbone and ResNet34\_vd as the text recognizer backbone, can achieve higher Hmean but slower inference speed than the ultra lightweight version. It can be found that the Hmean of PP-OCRv2 is 7.3\% higher than that of PP-OCR mobile models under the same inference cost, and is comparable to PP-OCR server models. Figure \ref{show_res} visualizes some end-to-end recognition results of the proposed PP-OCRv2 system and the previous ultra lightweight and large-scale PP-OCR system. 

\begin{table}[h]
\begin{center}
\begin{tabular}{c|c|c|c|c}
\hline
 & & & \multicolumn{2}{c}{Inference Time (ms)} \\
\cline{4-5}
\makecell[c]{Model \\ Type} & \makecell[c]{Hmean} & \makecell[c]{Model \\ Size (M)} & \makecell[c]{CPU} & \makecell[c]{T4 GPU}\\
\hline
\makecell[c]{PP-OCR  \\ mobile} & 0.503 & 8.1 & 356 & 116 \\ 

\makecell[c]{PP-OCR  \\ server} & 0.570 & 155.1 & 1056 & 200 \\

\makecell[c]{PP-OCRv2 \\ mobile} & 0.576 & 11.6 & 330 & 111 \\
\hline
\end{tabular}
\end{center}
\caption{Compare between PP-OCRv2 system and  PP-OCR mobile and server systems.}
\label{sys_scale}
\end{table}

\section{Conclusions}
In this paper, we proposed a more robust practical ultra lightweight OCR system PP-OCRv2. We introduced bag of tricks to enhance our previous work, PP-OCR, which include Collaborative Mutual Learning (CML),  CopyPaste, Lightweight CPU Network (PP-LCNet), Unified-Deep Mutual Learning (U-DML) and CenterLoss. Experiments on the real data show that the accuracy of the PP-OCRv2 is higher than PP-OCR at the same inference cost. The corresponding ablation experiments are also provided. Meanwhile, some practical ultra lightweight OCR models are released with a large-scale dataset.

\bibstyle{aaai21}
\bibliography{eg}

\end{document}